\documentclass{article}

\usepackage{microtype}
\usepackage{graphicx}
\usepackage{subcaption}
\usepackage{booktabs} 

\usepackage{hyperref}

\usepackage[preprint]{icml2026}

\usepackage{amsmath}
\usepackage{amssymb}
\usepackage{mathtools}
\usepackage{balance}
\usepackage{lastpage}
\usepackage{amsthm}

\usepackage[capitalize,noabbrev]{cleveref}

\theoremstyle{plain}

\theoremstyle{definition}

\theoremstyle{remark}

\usepackage[textsize=tiny]{todonotes}

\usepackage[nolist,nohyperlinks]{acronym}
\usepackage[acronym]{glossaries}
\begin{acronym}
\acro{CNN}{Convolutional Neural Networks}
\acro{SVM}{Support Vector Machine}
\end{acronym}

\begin{document}

\twocolumn[
  \icmltitle{Patch-Based Spatial Authorship Attribution in Human–Robot Collaborative Paintings}



  \icmlsetsymbol{equal}{*}

  \begin{icmlauthorlist}
    \icmlauthor{Eric Chen}{umich}
    \icmlauthor{Patricia Alves-Oliveira}{umich}
    
  \end{icmlauthorlist}

  \icmlaffiliation{umich}{Robotics, University of Michigan, Ann Arbor, Michigan, United States}

  \icmlcorrespondingauthor{Eric Chen}{erche@umich.edu}

  \icmlkeywords{Machine Learning, ICML}

  \vskip 0.3in
]



\printAffiliationsAndNotice{}  

\begin{abstract}
As agentic AI becomes increasingly involved in creative production, documenting authorship has become critical for artists, collectors, and legal contexts. We present a patch-based framework for spatial authorship attribution within human-robot collaborative painting practice, demonstrated through a forensic case study of one human artist and one robotic system across 15 abstract paintings. Using commodity flatbed scanners and leave-one-painting-out cross-validation, the approach achieves 88.8\% patch-level accuracy (86.7\% painting-level via majority vote), outperforming texture-based and pretrained-feature baselines (68.0\%--84.7\%). For collaborative artworks, where ground truth is inherently ambiguous, we use conditional Shannon entropy to quantify stylistic overlap; manually annotated hybrid regions exhibit 64\% higher uncertainty than pure paintings ($p$=0.003), suggesting the model detects mixed authorship rather than classification failure. The trained model is specific to this human-robot pair but provides a methodological grounding for sample-efficient attribution in data-scarce human-AI creative workflows that, in the future, has the potential to extend authorship attribution to any human-robot collaborative painting.
\end{abstract}

\section{Introduction}
Robotic and AI-assisted systems are increasingly participating in creative production, raising practical questions about authorship attribution. Attribution shapes valuation and trust in art markets. While the U.S. copyright doctrine emphasizes human authorship, when robotic systems produce physical paintings with brushstrokes visually similar to human work~\cite{deussen2012feedback, schaldenbrand2024cofrida}, distinguishing contributions becomes challenging. Contemporary practice increasingly includes collaborative workflows where humans and robots contribute to the same canvas, making authorship spatially varying—the evidentiary goal shifts from \textit{``who made this?''} to \textit{``where and how do contributions differ?''}

\begin{figure}[t]
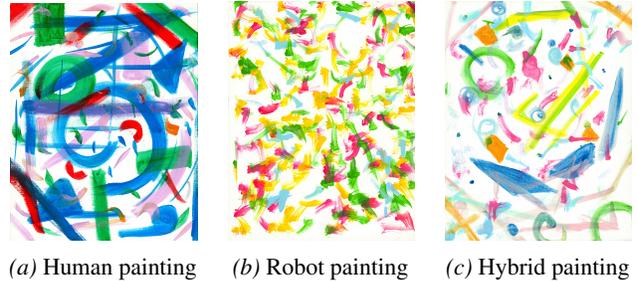

    \centering
    \begin{subfigure}{0.3\columnwidth}
        \centering
        \includegraphics[width=\linewidth]{Figures/artist_smaller.jpg}
        \caption{Human painting}
        \label{fig:human1}
    \end{subfigure}
    \hfill
    \begin{subfigure}{0.3\columnwidth}
        \centering
        \includegraphics[width=\linewidth]{Figures/robot_smaller.jpg}
        \caption{Robot painting}
        \label{fig:robot1}
    \end{subfigure}
    \hfill
    \begin{subfigure}{0.3\columnwidth}
        \centering
        \includegraphics[width=\linewidth]{Figures/humanrobot_smaller.jpg}
        \caption{Hybrid painting}
        \label{fig:hybrid1}
    \end{subfigure}
    
    
    
    \caption{Example of the paintings from the dataset showing human-created paintings (a), robot-created paintings generated using the CoFRIDA framework~\cite{schaldenbrand2024cofrida} (b), and collaborative human-robot paintings (c).}
    \label{fig:example_paintings}
\end{figure}

Prior computational approaches provide useful foundations but face constraints in these settings. Large-scale models achieve strong artist attribution with extensive corpora \citep{van2015toward}, but such datasets rarely exist for contemporary creators or robots with limited output. Scientific imaging approaches leverage fine-scale physical cues \citep{ji2021discerning} but require specialized instrumentation. We address these constraints with an approach compatible with small painting collections, commodity scanners, and spatial analysis within collaborative works.

We introduce a patch-based convolutional framework to address two key Research Questions (RQ): RQ 1. \textit{Can patch-level visual features distinguish human and robotic painting styles in data-scarce settings?} and RQ 2. \textit{Does model uncertainty provide a principled signal for identifying regions of mixed 
authorship in collaborative artworks?} 

High-resolution scans are partitioned into patches classified as blank canvas, human-created, or robot-created based on learned brushstroke representations, thereby enabling spatial attribution with commodity flatbed scanners. Using 15 paintings (7 human, 8 robot) and leave-one-painting-out cross-validation, we achieve 88.8\% patch-level accuracy (86.7\% painting-level via majority vote), demonstrating the model learns consistent stylistic differences rather than painting-specific artifacts. For collaborative artworks in which patch-level ground truth is inherently ambiguous, we use predictive entropy to identify regions of mixed authorship: patches from collaborative works exhibit 64\% higher uncertainty than those in pure paintings ($p=0.003$), suggesting that the model detects overlapping stylistic cues characteristic of both agents.

\section{Related Works}
We frame our contribution within prior efforts on human-produced art and authentication, and on advances in robotic and AI-generated art.

\subsection{Human Produced Art and Authentication}
\subsubsection{Art Authentication and Authorship}
Art authentication—the scholarly determination of an artwork's creator through systematic examination of stylistic characteristics, material evidence, and documentary provenance~\cite{spencer2004expert}—has traditionally relied on expert connoisseurship, provenance documentation, and scientific examination of materials. Brushstrokes serve as a defining component of artistic style, reflecting individual motor behavior~\cite{arnheim1954art}. Computational analysis has explored vectorized stroke representations~\cite{hertzmann1998painterly}, stroke-level feature modeling~\cite{li2011rhythmic}, and brushwork dynamics~\cite{salisbury1994interactive}. As robotic and AI systems increasingly participate in creative production~\cite{davis2016empirically, schaldenbrand2024cofrida}, authorship attribution in collaborative workflows becomes especially salient. However, existing methods assume single authorship, making them unsuitable for collaborative pieces where authorship varies spatially.

Texture-based approaches have also been applied to art authentication and style classification. Local texture descriptors such as Local Binary Patterns and Gabor filters~\cite{shamir2010impressionism} have been shown to capture stylistic information in paintings. Wavelet-based methods~\cite{johnson2008image} further introduced multi-resolution representations for painter identification. Physical authentication methods also analyze three-dimensional surface topography~\cite{ji2021discerning}, crack patterns~\cite{sizyakin2020crack}, and material composition via X-ray or infrared imaging~\cite{goodall2012identification}. These approaches provide strong discriminative signals but require specialized equipment, limiting accessibility for documenting contemporary collaborative practices.

\subsubsection{Computational Authentication}
Computational approaches to art authentication aim to recover stylistic and process-level signals from visual evidence. Large-scale convolutional models trained on museum collections achieve strong artist attribution~\cite{karayev2013recognizing, van2015toward}, but require thousands of paintings per artist and assume single authorship. Commercial systems such as Art Recognition~\cite{artrecognition2024} employ deep learning for the authentication of established artists, but depend on extensive catalogs of authenticated works typically unavailable for contemporary creators, emerging artists, or robotic systems with limited output. Localized methods provide complementary analysis: stroke-level features demonstrate robustness to forgeries~\cite{elgammal2018picasso}, while fine-grained hatching patterns can distinguish artists~\cite{ziaee2025fine}.

Patch-based learning enables spatial analysis when discriminative information is localized~\cite{hou2016patch, xu2014deep}, increasing effective dataset size by treating local regions as training samples. In the art domain, patch-based methods have been applied to style analysis and forgery detection~\cite{ji2021discerning, azimi2024patch, van2025patch}, though existing approaches assume all patches share the same author label. Physical property analysis via surface topography~\cite{ji2021discerning} and crack patterns~\cite{sizyakin2020crack} provides complementary evidence but requires specialized equipment. In contrast, our approach operates in data-scarce settings without extensive reference catalogs, uses commodity scanning hardware, and explicitly considers spatially varying authorship within collaborative artworks.

\subsection{Robotic and AI-Generated Art}
\subsubsection{Robotic Art Systems}
Robotic painting systems translate perceptual goals into physical paint application through iterative perception–action loops. Representative systems include e-David, which uses visual feedback to guide stroke placement~\cite{deussen2012feedback}, robotic portrait-drawing platforms such as Paul the Robot~\cite{tresset2013portrait}, and collaborative systems such as FRIDA and CoFRIDA~\cite{schaldenbrand2022frida, schaldenbrand2024cofrida}. More recent systems emphasize adaptability and interaction, enabling robots to respond to partially completed canvases or human input during the creative process.

These systems demonstrate that robots can exhibit consistent, repeatable painting behaviors arising from control strategies, perception pipelines, and mechanical constraints. As a result, robotic paintings may exhibit systematic visual signatures distinct from those of human artists, despite operating within shared materials and tools.

\begin{figure}[t]
    \centering
    \includegraphics[width=0.40\textwidth]{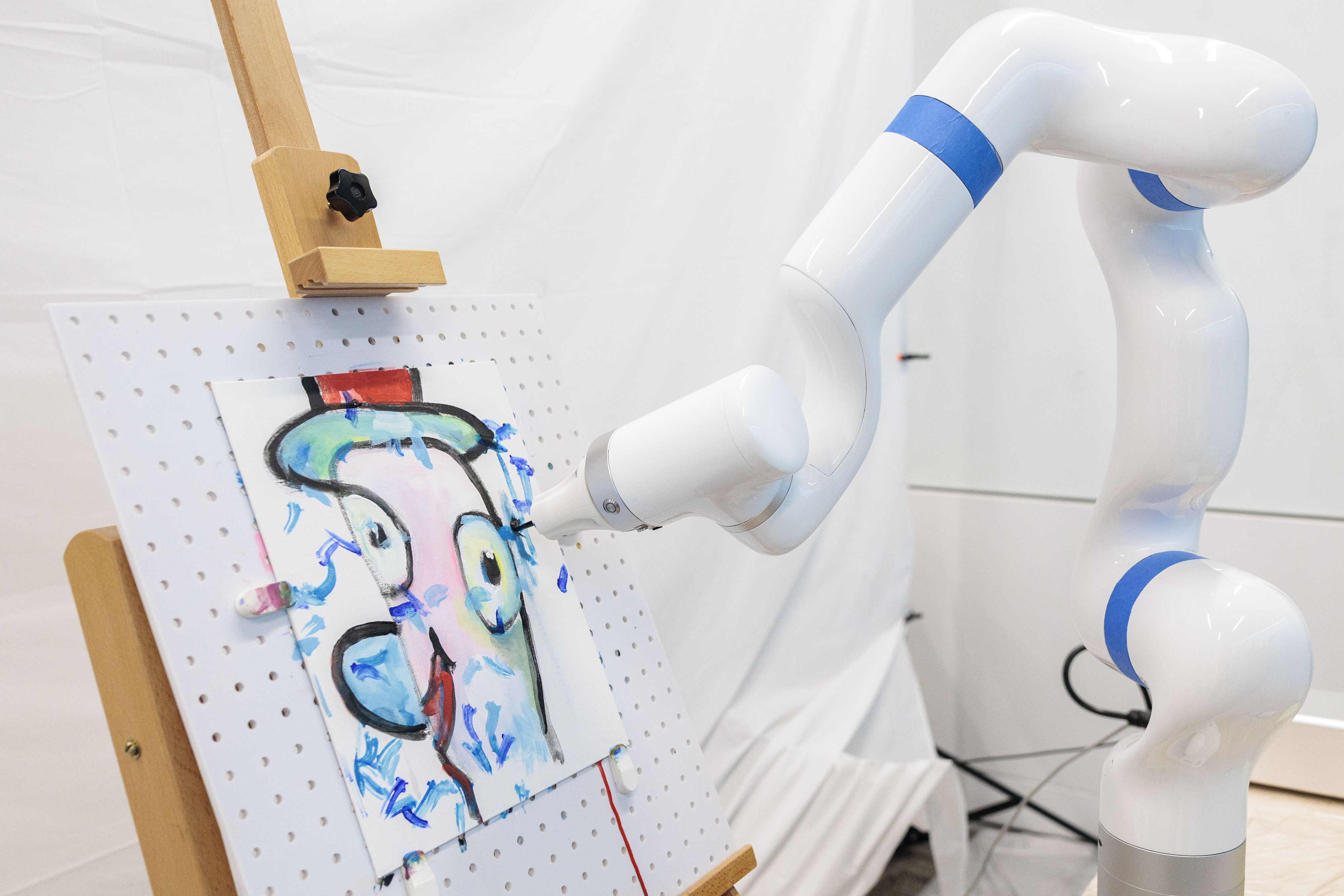}
    \caption{The robotic painting system in a collaborative painting session, illustrating the workflow used to generate hybrid paintings.}
    \label{fig:robot_painter}
\end{figure}

\subsubsection{Authentication of Robotic and AI-Generated Art}
The rise of AI-generated and robotic art motivates growing interest in questions of authorship, originality, and authenticity~\cite{zylinska2020ai}. Most technical work focuses on detecting digitally generated or manipulated images, exploiting artifacts introduced by neural rendering, synthesis pipelines, or compression~\cite{rossler2019faceforensics++, verdoliva2020media}. These methods are effective for digital content but do not transfer to physical paintings, where visual characteristics emerge through embodied interaction with materials and are later digitized via scanning or photography.

Human studies on identifying AI-generated art suggest that perceptual discrimination is unreliable and context-dependent, particularly as generative models improve. In the case of robotic painting, authentication becomes even more challenging: the final artifact may closely resemble human-produced work, while underlying differences arise from process rather than appearance alone.

\section{Dataset Construction}

\subsection{Paintings and Digitization}
Our dataset consisted of 15 physical paintings: 7 created by a single human artist using acrylic on canvas, and 8 generated by the robotic system described in~\ref{sec:robot_paintings}. All paintings were produced using the same canvas type, acrylic paint, and brush materials, with identical scanning settings. We deliberately limit our study to a single human-robot pair for two reasons. First, this reflects realistic deployment scenarios where an artist documents their own collaborative practice and lacks access to large multi-artist datasets. Second, it allows us to focus on methodological development, demonstrating that spatial attribution is possible with minimal data—rather than making premature claims about universal human vs robot stylistic differences.

All paintings were digitized using commodity flatbed scanning at 1200 DPI. This controlled acquisition reduces confounding factors such as lighting variation and camera-specific artifacts that provide spurious classification cues.

\subsection{Patch Extraction and Statistical Units}
Each painting was divided into 300×300 pixel patches with 50\% overlap (stride 150), yielding 136{,}977 patches. Patches were labeled as \textit{Blank} ($\geq$95\% white), \textit{Human}, or \textit{Robot} based on source painting. Dense extraction creates spatial correlation; thus, the painting is the statistical unit, not the patch. All validation uses leave-one-painting-out cross-validation~\cite{hastie2009elements} (Section ~\ref{sec: crossval}), ensuring no patches from held-out paintings appear during training. This reflects deployment where models analyze new paintings.

\subsection{Human Paintings} \label{sec:human_paintings}
The seven human-painted canvases were created by a single artist using acrylic on canvas. The artist was given the thematic prompt ``air'' as an abstract conceptual starting point, allowing stylistic and compositional divergence while maintaining a shared thematic foundation. This loose constraint provided sufficient creative freedom to generate diverse visual characteristics while avoiding arbitrary subject matter. The resulting paintings exhibit variation in brushstroke density, color palette, and compositional structure, reflecting individual artistic choices within the thematic framework.

\subsection{Robot Paintings} \label{sec:robot_paintings}

Our eight robot-painted canvases were generated using the CoFRIDA framework~\cite{schaldenbrand2022frida,schaldenbrand2024cofrida} executed on a 6-DOF UF850 robotic arm equipped with a paintbrush. CoFRIDA follows an image-to-action pipeline in which abstract target images are first generated using a fine-tuned InstructPix2Pix model with randomly sampled text prompts. These target images represent high-level visual goals for a painting session. Given a target image, FRIDA computes a sequence of brush strokes that iteratively approximates the goal and executes them on the canvas to produce the final painting.

\subsection{Hybrid Paintings} \label{sec:hybrid_paintings}
The five hybrid paintings were produced in a collaborative setting in which the robot executed pre-generated stroke sequences while the human artist painted concurrently on the same canvas (Figure~\ref{fig:robot_painter}). The robot followed the same randomly sampled 
prompt-based generation process as the pure robot paintings, while the human artist responded to the emerging composition. These mixed canvases provide combined human and robot contributions and are used to evaluate patch-level classification in collaborative painting scenarios.


\section{Model Architecture}
We use a VGG-style convolutional neural network adapted from the PigeoNET framework~\cite{van2015toward}. VGG architectures have 
demonstrated strong performance on texture recognition 
tasks~\cite{cimpoi2015deep}, while patch-based methods enable 
effective learning in data-scarce settings by treating local 
regions as training samples~\cite{hou2016patch, xu2014deep}.

The network processes grayscale 300×300 pixel patches through five convolutional blocks (32→64→128→256→512 filters), each containing two 3×3 convolutional layers with batch normalization, ReLU activation, and 2×2 max pooling. After the final block, dropout ($p$=0.4), global average pooling, and three fully connected layers (512→512→128→3) produce logits for blank, human, and robot classes.

We use a compact architecture rather than fine-tuning large pretrained models because our setting (14 training paintings per fold) provides limited supervision and risks overfitting, and because task-specific learning can better capture fine-grained brushstroke cues relevant to forensic attribution.


\section{Training and Optimization}

\subsection{Data Augmentation}
We applied standard augmentations: rotation (±15°), flips, random resized cropping (0.8–1.0), Gaussian blur (kernel 3), and random cropping with padding (10px).

\subsubsection{Class Imbalance Handling}
Our dataset exhibits significant class imbalance, with substantially more human patches than robot or blank patches. To address this, we employed a weighted cross-entropy loss function with class-specific weights. Weights were computed as 
\begin{equation} 
    w_c = \alpha_c × (N_{total} / N_c)
\end{equation}

, where $\alpha_c$ is a scaling factor set to 0.01 for blank patches (to downweight this trivially distinguishable class), 1.0 for human patches (baseline), and 0.75 for robot patches (to maintain balanced emphasis on both primary classes without overprioritizing the minority class). 

\subsection{Optimization}
We trained all models using stochastic gradient descent (SGD) with learning rate 0.0001 and momentum 0.9. Training used batch size 64 for 100 epochs. Models were implemented in PyTorch and trained on an NVIDIA GeForce RTX 4090 GPU. 

\section{Experimental Evaluation}
This section explains how we evaluated the performance of our approach. 

\subsection{Cross Validation on Pure Paintings} \label{sec: crossval}
To evaluate generalization while avoiding overfitting to individual artworks, we employed a leave-one-painting-out cross-validation~\cite{hastie2009elements}. In each of the 15 folds, where a fold represents one complete train-test split 
with a different held out painting, a single painting was held out entirely for validation, while the model was trained on patches extracted from the remaining paintings. Crucially, no patches from the held-out painting were seen during training, ensuring that the evaluation was performed on completely unseen artworks rather than visually similar patches from the same painting. We hold out entire paintings (not patches) to prevent leakage from spatially correlated regions and to evaluate generalization to unseen artworks.


For each fold, model training proceeds until convergence, and we select the checkpoint achieving the highest validation accuracy on the held-out painting. The selected model is then used to compute all reported metrics for that fold. Performance is aggregated across folds by reporting the mean and standard deviation, providing insight into both average accuracy and variability across different held-out artworks.

\subsection{Evaluation Metrics}
For the leave-one-painting-out cross-validation experiments, we report patch-level classification performance using the following metrics:

\begin{itemize}\itemsep=0em
    \item Overall accuracy: Percentage of correctly classified patches across all categories;
    \item Per-class accuracy: Classification accuracy for \textit{Blank}, \textit{Human}, and \textit{Robot} patches separately;
    \item Precision and recall: Per-class precision and recall to assess performance on imbalanced categories;
    \item Confusion matrices: To analyze specific misclassification patterns (e.g., human patches misclassified as robot).
\end{itemize}

All metrics are computed independently for each fold and summarized by reporting the mean and standard deviation across folds, providing insight into both average performance and variability across held-out paintings.  

\subsection{Baseline Methods}




To contextualize the proposed patch-based CNN, we compare against three baseline approaches representing commonly used alternatives in art analysis and visual classification. These baselines are not intended to exhaustively evaluate architectural choices, but rather to assess whether the proposed formulation provides benefits beyond established feature representations.

The first baseline used hand-crafted texture descriptors. We extracted Local Binary Pattern (LBP) features~\cite{ojala1994performance} from each patch and trained a Random Forest classifier~\cite{breiman2001random}, following prior work on texture-based artist identification. The second and third baselines used pretrained deep features from ResNet-50~\cite{he2016deep} and DINOv2~\cite{oquab2023dinov2} models pretrained on ImageNet, training a linear \ac{SVM} classifier~\cite{cortes1995support} on frozen representations. ResNet-50 represents traditional supervised pretraining, while DINOv2 employs self-supervised learning on diverse visual data. All baselines were evaluated using the same leave-one-painting-out cross-validation protocol.

\subsection{Analysis of Hybrid Human-Robot Paintings}

While quantitative evaluation is feasible for pure human-only and robot-only paintings, hybrid human-robot artworks present a fundamentally different challenge. In these collaborative paintings, human and robotic brushstrokes may overlap spatially, interleave temporally, or influence one another stylistically. As a result, precise patch-level ground truth attribution is often ill-defined and ambiguous even for human annotators. Consequently, we did not evaluate hybrid paintings using standard classification accuracy or error-based metrics. 

We analyzed hybrid paintings through the lens of \textit{model uncertainty}, using predictive entropy as a proxy for stylistic ambiguity between human and robot contributions. Our model was trained exclusively on patches from pure paintings where authorship is unambiguous. When applied to hybrid paintings, the model interprets patches that may exhibit mixed or intermediate stylistic characteristics. We hypothesize that patches with genuinely mixed authorship will elicit higher uncertainty than patches from pure paintings, providing a principled signal for identifying collaborative regions without requiring ground truth labels.

\subsubsection{Conditional Entropy Formulation} \label{sec: cond entr}

For each patch, the model produces a probability distribution $\mathbf{p} = [p_{\text{blank}}, p_{\text{human}}, p_{\text{robot}}]$ over three classes. To isolate ambiguity specifically between human and robot authorship, we computed the Shannon entropy~\cite{shannon1948mathematical} of the renormalized human-robot posterior, conditioned on the patch containing painted content. 

We used conditional Shannon entropy rather than alternative uncertainty measures for three reasons: (1) it is invariant to class imbalance after renormalization; (2) it has an interpretable upper bound (1 bit) independent of model calibration; and (3) it directly quantifies stylistic overlap in learned representations rather than epistemic uncertainty about an unknown ground truth. Entropy elevation in hybrid paintings indicates the model detects features characteristic of both agents, which it learned to distinguish when presented in isolation.

Formally, for patches where $p_{\text{human}} + p_{\text{robot}} > \tau$ ($\tau = 0.2$), we compute:
\begin{align}
P_{\text{human}} &= \frac{p_{\text{human}}}{p_{\text{human}} + p_{\text{robot}}}, \\
P_{\text{robot}} &= \frac{p_{\text{robot}}}{p_{\text{human}} + p_{\text{robot}}}, \\
H &= -P_{\text{human}} \log_2(P_{\text{human}}) - P_{\text{robot}} \log_2(P_{\text{robot}}).
\end{align}

This conditional entropy $H \in [0, 1]$ ranges from zero (high confidence attribution to either human or robot) to one bit (maximal ambiguity, corresponding to $P_{\text{human}} = P_{\text{robot}} = 0.5$). We selected ($\tau$) to exclude low-content regions; exploring sensitivity to this threshold is left to future work.

By conditioning on non-blank confidence ($\tau$), this measure avoids confounding stylistic ambiguity with uncertainty arising from empty or sparsely painted regions of the canvas. High entropy values therefore indicate patches where the model cannot confidently determine whether the observed visual characteristics are more consistent with human or robotic painting styles.

\subsubsection{Manual Annotation Protocol}

We manually annotated hybrid regions in collaborative paintings by visually identifying patches where both human and robot contributions appeared to be present. This process involved examining high-resolution scans and selecting 300$\times$300 pixel regions exhibiting one or more of the following characteristics: (1) visible overlapping of human and robot brushstrokes; (2) interleaved strokes from both agents within close proximity; or (3) regions where paint application from one agent appeared to modify or interact with strokes from the other agent.

These annotations were not intended to establish definitive patch-level authorship. Instead, they identified regions of potential collaborative contribution, enabling targeted analysis of model behavior in areas where mixed authorship is expected. This framing motivates our use of uncertainty-based analysis rather than accuracy metrics, and allowed direct comparison between hybrid regions and pure human or robot paintings.

\section{Results}

\subsection{Overall Cross-Validation Performance}
We evaluated our model using 15-fold leave-one-painting-out cross-validation on the dataset comprising 7 human paintings and 8 robot paintings. Table~\ref{tab:cv_results} presents the overall performance metrics averaged across all folds.

\begin{table}[t]
\centering
\caption{Comparison of baseline methods and the proposed approach using 15-fold leave-one-painting-out cross-validation.}
\label{tab:baselines}
\small
\begin{tabular}{lcc}
\toprule
\textbf{Method} & \textbf{Mean Acc.} & \textbf{Std.} \\
\midrule
LBP + RF & 65.92\% & 20.90\% \\
ResNet-50 + SVM & 81.96\%  & 10.76\% \\
DINOv2 + SVM & 84.70\% & 10.05\% \\
\textbf{Ours (CNN)} & \textbf{88.79\%} & \textbf{10.41\%} \\
\bottomrule
\end{tabular}
\end{table}

Table~\ref{tab:baselines} compares baseline methods and our approach. Hand-crafted texture features (LBP + Random Forest) achieve 65.92\% with high variance, indicating limited robustness. Pretrained deep features improve considerably: DINOv2 + SVM achieves 84.70\% and ResNet-50 + SVM achieves 81.96\%. While these generic representations capture stylistic information despite the domain mismatch between natural images and abstract paintings, frozen features cannot adapt to specific visual signatures of this human-robot pair.

The proposed patch-based CNN outperforms baselines at 88.79\% accuracy. The 4\% improvement over DINOv2 indicates that learning directly from painting patches captures fine-grained brushstroke characteristics not emphasized in natural image pretraining. Critically, our approach achieves these gains without large-scale pretraining, using a compact architecture with orders of magnitude fewer parameters than foundation models, enhancing interpretability and computational efficiency 
for data-scarce, per-artist authentication.

\begin{table}[h]
\centering
\caption{Cross-validation results across 15 folds with per-class performance metrics.}
\label{tab:cv_results}
\begin{tabular}{lcc}
\toprule
\textbf{Metric} & \textbf{Mean} & \textbf{Std Dev} \\
\midrule
Overall (Micro-averaged) Accuracy & 88.79\% & 10.41\% \\
Balanced Accuracy & 87.86\% \\
\midrule
\multicolumn{3}{c}{\textit{Per-Class Recall (Patch-Level)}} \\
Blank (11 folds) & 85.01\% & 14.12\% \\
Human (7 folds) & 88.33\% & 12.05\% \\
Robot (8 folds) & 90.24\% & 8.42\% \\
\midrule
Total Validation Patches & \multicolumn{2}{c}{136,977} \\
\quad Blank & \multicolumn{2}{c}{17,553} \\
\quad Human & \multicolumn{2}{c}{60,217} \\
\quad Robot & \multicolumn{2}{c}{59,207} \\
\bottomrule
\end{tabular}
\end{table}

\subsection{Performance Analysis}
Per-class analysis reveals strong performance across all three categories. The model achieved 90.24\% accuracy on robot patches, 88.33\% on human patches, and 85.01\% on blank canvas patches. Performance on blank patches is slightly lower than for painted regions, reflecting the inherent ambiguity of patches near transitions between unpainted canvas and very light or sparse brushstrokes, which occur frequently in paintings with large unpainted areas. The high robot classification accuracy (90.24\%) demonstrates that the model successfully learns distinctive textural features of this robotic system's painting despite robot paintings also containing blank regions. 

Performance varied across paintings, with accuracies ranging from 69.0\% to 99.4\% (Appendix Table~\ref{tab:per_fold_appendix}). Lower accuracies occurred primarily on paintings with substantial blank regions, creating complex boundary transitions between painted and unpainted areas.

\begin{figure*}[t]
    \centering
    \begin{subfigure}{0.45\textwidth}
      \centering
      \includegraphics[height=6.5cm]{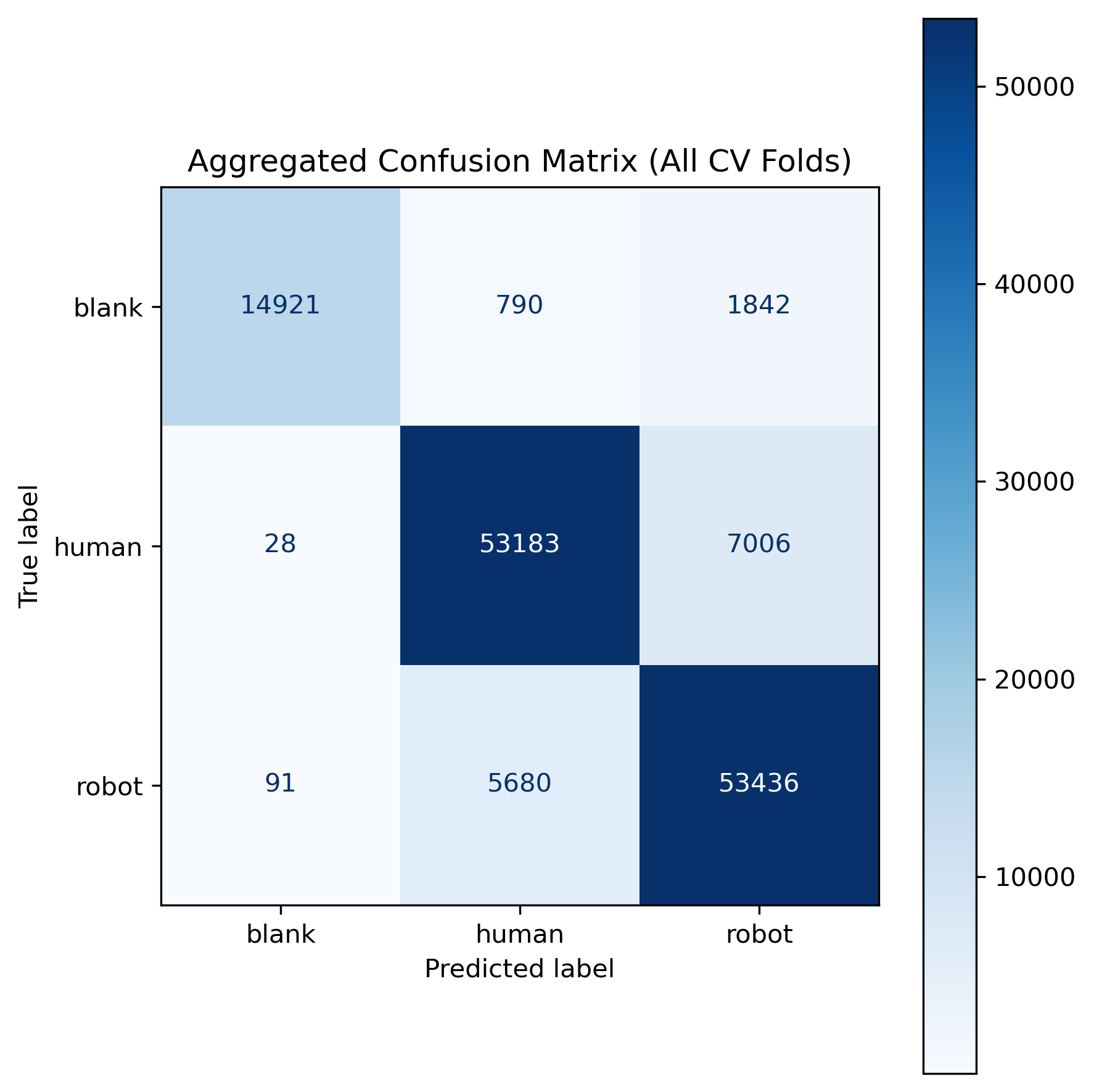}
      \caption{Aggregated confusion matrix (raw counts).}
      \label{fig:1a}
    \end{subfigure}
    \hspace{0.03\textwidth}
    \begin{subfigure}{0.45\textwidth}
      \centering
      \includegraphics[height=6.5cm]{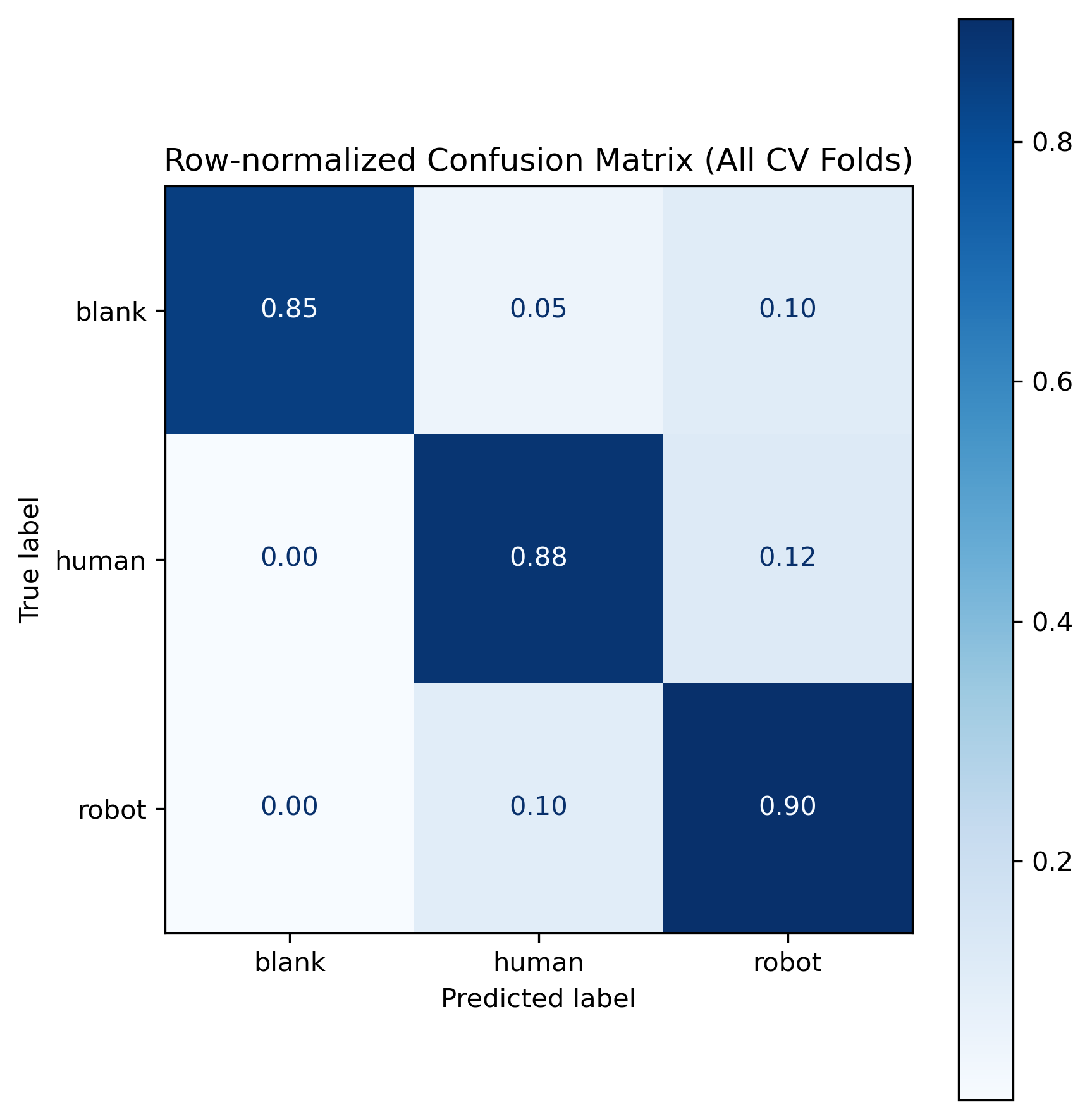}
      \caption{Row-normalized confusion matrix (per-class recall).}
      \label{fig:1b}
    \end{subfigure}
    \caption{Patch-level confusion matrices aggregated across all cross-validation folds. (a) Raw counts. (b) Per-class recall.}
    \label{fig:confusion_matrices}
\end{figure*}

\subsection{Confusion Matrix Analysis}

Figure~\ref{fig:confusion_matrices} presents patch-level confusion matrices aggregated across all cross-validation folds, illustrating both absolute error distributions and per-class behavior. The raw count matrix shows that most misclassifications occur between human and robot patches, reflecting the visual similarity between human and robot-generated strokes in the abstract painting domain. Blank patches are generally well separated, with relatively few errors, though a small fraction are misclassified as robot patches, likely due to low-texture regions or sparse stroke artifacts.

The row-normalized confusion matrix further highlights per-class recall independent of class imbalance. All three classes achieve strong recall (approximately 85–90\%), indicating consistent performance across painting categories despite differences in patch frequency. The dominant failure mode remains human–robot confusion, suggesting that while the model captures broad stylistic differences, fine-grained distinctions between human and robot stroke patterns remain challenging at the patch level. Overall, these results demonstrate that the classifier learns meaningful stylistic cues while maintaining balanced performance across classes.

\subsection{Hybrid Human-Robot Painting Analysis}
To investigate authorship uncertainty in collaborative settings, we analyzed predictive entropy distributions across pure and hybrid paintings. For controlled comparison, we randomly selected 5 human paintings and 5 robot paintings from our dataset to match the sample size of 5 available hybrid paintings. This balanced design (5 paintings per category) enables direct statistical comparison of entropy distributions while accounting for per-painting variability.

Pure painting entropy statistics were computed using leave-one-painting-out cross-validation to ensure the model's predictions on each pure painting reflected genuinely unseen data, mirroring the deployment scenario for hybrid paintings. Hybrid painting statistics were computed from 1,640 manually annotated patches across 5 collaborative artworks, representing regions where both human and robot contributions were visually identifiable.

\begin{table}[ht]
\centering
\caption{Predictive entropy comparison between pure and hybrid paintings. Pure painting statistics computed via leave-one-painting-out cross-validation on 5 randomly selected paintings per class. Hybrid statistics computed from 1{,}640 manually annotated patches across 5 collaborative paintings.}
\label{tab:entropy_comparison}
\small
\begin{tabular}{lccc}
\toprule
\textbf{Metric} & \textbf{Pure Human} & \textbf{Pure Robot} & \textbf{Hybrid} \\
\midrule
n patches & 44{,}771 & 37{,}059 & 1{,}640 \\
n paintings & 5 (CV) & 5 (CV) & 5 \\
\midrule
Median & 0.11 & 0.13 & \textbf{0.18} \\
Mean & 0.22 & 0.20 & \textbf{0.34} \\
Std Dev & 0.28 & 0.27 & 0.35 \\
IQR & 0.04--0.33 & 0.04--0.29 & 0.03--0.65 \\
\midrule
\% $H > 0.5$ & 17.0\% & 15.0\% & \textbf{33.1\%} \\
\% $H > 0.7$ & 10.3\% & 9.6\% & \textbf{23.4\%} \\
\% $H > 0.9$ & 5.1\% & 4.8\% & \textbf{11.1\%} \\
\midrule
Mean of medians & 0.12 $\pm$ 0.11 & 0.11 $\pm$ 0.11 & \textbf{0.19 $\pm$ 0.06} \\
\bottomrule
\end{tabular}

\vspace{0.5em}
\footnotesize
\textit{Statistical test:} Mann-Whitney U test on painting-level medians: Pure vs Hybrid U=0, p=0.003; Pure Human vs Pure Robot U=12, p=1.0.
\end{table}

Table~\ref{tab:entropy_comparison} presents entropy statistics for pure and hybrid paintings. Hybrid patches exhibited systematically elevated entropy compared to held-out pure paintings, with pooled median entropy of 0.18 representing a 1.5× increase over pure paintings (medians 0.11-0.13, Mann-Whitney U=0, p=0.003). Individual hybrid paintings showed median entropy ranging from 0.11 to 0.27, demonstrating consistent elevation across five independent collaborative artworks despite variation in collaboration style.

Notably, pure human and pure robot paintings exhibited nearly identical entropy distributions (medians 0.11 vs 0.13), confirming that elevation in hybrid patches reflects mixed authorship cues rather than general model uncertainty or domain shift. Statistical comparison confirms this symmetry: Mann-Whitney U test between pure human and pure robot painting medians yields U=12, p=1.0, indicating no significant difference in entropy distributions between the two pure categories. This symmetry validates that the model learned genuine stylistic differences between human and robot painting rather than artifacts of image acquisition or canvas properties.

\begin{figure*}[t]
    \centering
    \includegraphics[width=0.85\textwidth]{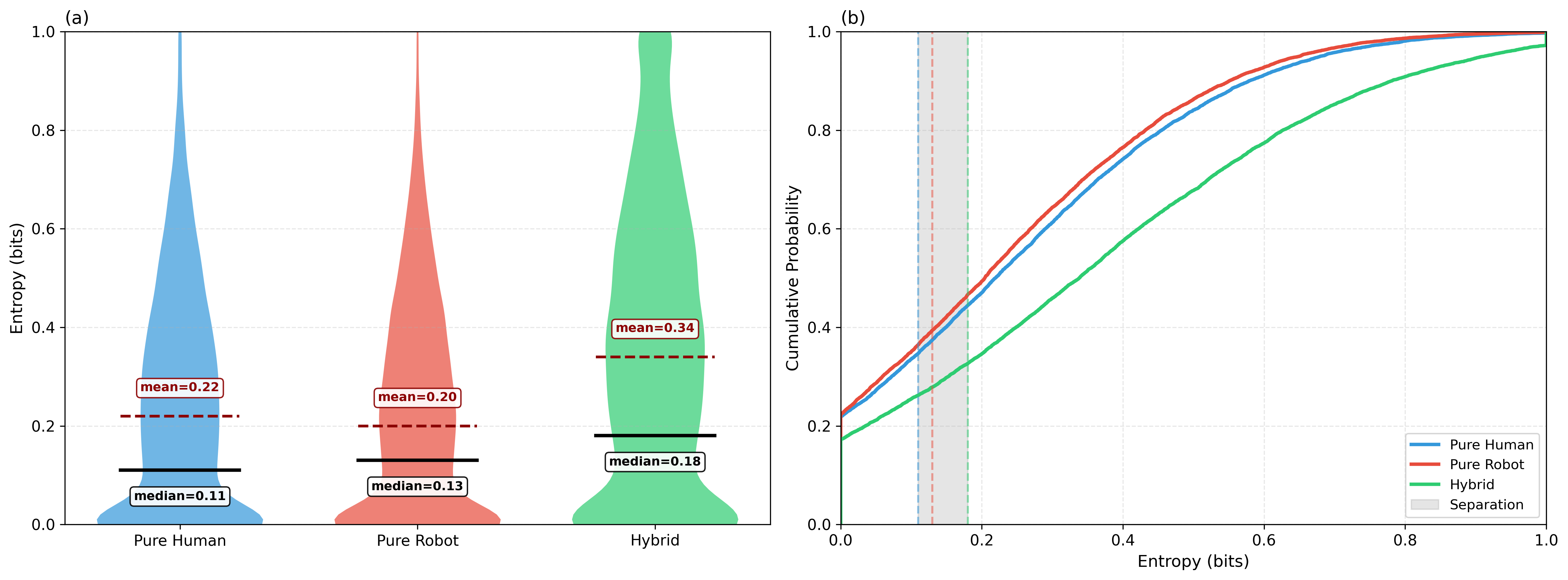}
    \caption{Entropy analysis comparing pure and hybrid paintings. (a) Violin plots show pure human and robot paintings exhibit similar low-entropy distributions (medians 0.11 and 0.13), while hybrid paintings display elevated uncertainty (median 0.18). (b) CDFs demonstrate systematic separation between pure and hybrid paintings. Dashed lines indicate median entropy for each category.}
    \label{fig:entropy_analysis}
\end{figure*}

Figure~\ref{fig:entropy_analysis} visualizes the entropy distributions for pure human, pure robot, and hybrid paintings described in Table~\ref{tab:entropy_comparison}. The violin plots (a) reveal the similarity between pure categories and the elevated spread in hybrid paintings. The cumulative distribution functions (b) demonstrate that this elevation is not driven by outliers but represents a systematic shift: approximately 50\% of pure painting patches exhibit entropy below 0.12, while the same proportion of hybrid patches exceed 0.18. The minimal overlap between pure and hybrid CDFs indicates that predictive uncertainty provides a robust signal of collaborative authorship.

Across the five collaborative paintings, 33\% of hybrid patches exceeded $H=0.5$ (vs. 16\% for pure), 23\% exceeded $H=0.7$ (vs. 10\%), and 11\% approached maximal uncertainty at $H>0.9$ (vs. 5\%). This 2× increase in high-uncertainty patches, consistent across all independent artworks, indicates that collaborative authorship produces patches where human and robot attribution cues are ambiguous. The model's uncertainty reflects its detection of features characteristic of both agents, which it learned to distinguish when present in isolation in pure paintings.

\subsection{Validation of Patch-Based Approach}

To verify that the patch-based formulation provides meaningful sample efficiency rather than merely inflating evaluation statistics, we compare performance under three complementary evaluation regimes.

First, we report standard patch-level classification accuracy under leave-one-painting-out cross-validation, achieving 88.8\%. Second, we perform majority-vote aggregation over patches: for each held-out painting, we assign a single predicted label based on the most frequent non-blank patch prediction. Under this regime, the model correctly classifies 13 out of 15 paintings (86.7\% painting-level accuracy), indicating patch-level predictions are internally consistent and reflect stable painting-level stylistic differences. Third, we evaluate a constrained baseline where a single patch is randomly sampled from each painting for training and testing, yielding one sample per painting. Performance drops substantially to 62.3\% $\pm$ 8.1\% (mean $\pm$ standard deviation over 10 random seeds), approaching chance-level.

These results demonstrate that patch-based learning provides meaningful representational and statistical benefits beyond pseudoreplication, enabling the model to learn consistent painting-level stylistic structure from a small number of physical artworks.

\section{Conclusion and Future Work}
Computational art authentication has evolved from large-scale CNN models~\cite{karayev2013recognizing, van2015toward} and stroke-level analysis~\cite{elgammal2018picasso, ziaee2025fine} to commercial systems like Art Recognition~\cite{artrecognition2024}. However, existing approaches assume single authorship and require extensive reference catalogs. Patch-based methods~\cite{ji2021discerning, azimi2024patch, van2025patch} provide spatial analysis but similarly assume homogeneous authorship. Our work addresses these limitations by operating with minimal data, using commodity hardware, and explicitly modeling spatially varying authorship in collaborative works.

We presented a patch-based framework for spatial authorship attribution in human–robot collaborative painting under data scarcity, demonstrated on a single human–robot pair. Using leave-one-painting-out cross-validation, we achieved 88.8\% patch-level accuracy (86.7\% painting-level via majority vote), outperforming texture-based and pretrained-feature baselines. For collaborative artworks where patch-level ground truth is inherently ambiguous, we used predictive entropy to identify regions of stylistic overlap where definitive attribution is ill-defined. Manually annotated hybrid regions exhibited 64\% higher uncertainty than pure paintings ($p$=0.003), demonstrating that model uncertainty reflects mixed authorship rather than classification failure. 

Future work includes extending the framework to multiple human artists and robotic systems to study the generality of learned stylistic cues, evaluating cross-system transfer between different human–robot pairs, and incorporating temporal or process-level information such as stroke ordering and robot motion trajectories. These directions build directly on the uncertainty-aware, patch-based methodology established here.

\section*{Impact Statement}

This paper presents work whose goal is to advance the field of Machine Learning. Our work on distinguishing human and robotic painting styles could have applications in art authentication and attribution. As with any computational attribution system, potential impacts include: (1) overconfidence in automated authentication, which should complement rather than replace expert analysis, (2) misuse in disputes over authorship or attribution, and (3) privacy concerns if applied without consent. We emphasize that our approach provides evidence to support human judgment, not definitive attribution claims. 

\section*{Acknowledgments}
This was supported by DARPA \#D24AP00323.


\balance
\bibliography{references}

@article{van2015toward,
  title={Toward Discovery of the Artist's Style: Learning to recognize artists by their artworks},
  author={Van Noord, Nanne and Hendriks, Ella and Postma, Eric},
  journal={IEEE Signal Processing Magazine},
  volume={32},
  number={4},
  pages={46--54},
  year={2015},
  publisher={IEEE}
}

@article{ji2021discerning,
  title={Discerning the painter’s hand: machine learning on surface topography},
  author={Ji, Fang and McMaster, Michael S and Schwab, Samuel and Singh, Gundeep and Smith, Lauryn N and Adhikari, Shishir and O’Dwyer, M{\'a}rcio and Sayed, Farah and Ingrisano, Anthony and Yoder, Dean and others},
  journal={Heritage Science},
  volume={9},
  number={1},
  pages={1--11},
  year={2021},
  publisher={Nature Publishing Group}
}

@book{arnheim1954art,
  title={Art and visual perception: A psychology of the creative eye},
  author={Arnheim, Rudolf},
  year={1954},
  publisher={Univ of California Press}
}

@inproceedings{hertzmann1998painterly,
  title={Painterly rendering with curved brush strokes of multiple sizes},
  author={Hertzmann, Aaron},
  booktitle={Proceedings of the 25th annual conference on Computer graphics and interactive techniques},
  pages={453--460},
  year={1998}
}

@article{li2011rhythmic,
  title={Rhythmic brushstrokes distinguish van Gogh from his contemporaries: findings via automated brushstroke extraction},
  author={Li, Jia and Yao, Lei and Hendriks, Ella and Wang, James Z},
  journal={IEEE transactions on pattern analysis and machine intelligence},
  volume={34},
  number={6},
  pages={1159--1176},
  year={2011},
  publisher={IEEE}
}

@inproceedings{salisbury1994interactive,
  title={Interactive pen-and-ink illustration},
  author={Salisbury, Michael P and Anderson, Sean E and Barzel, Ronen and Salesin, David H},
  booktitle={Proceedings of the 21st annual conference on Computer graphics and interactive techniques},
  pages={101--108},
  year={1994}
}

@article{shamir2010impressionism,
  title={Impressionism, expressionism, surrealism: Automated recognition of painters and schools of art},
  author={Shamir, Lior and Macura, Tomasz and Orlov, Nikita and Eckley, D Mark and Goldberg, Ilya G},
  journal={ACM Transactions on Applied Perception (TAP)},
  volume={7},
  number={2},
  pages={1--17},
  year={2010},
  publisher={ACM New York, NY, USA}
}

@article{johnson2008image,
  title={Image processing for artist identification},
  author={Johnson, C Richard and Hendriks, Ella and Berezhnoy, Igor J and Brevdo, Eugene and Hughes, Shannon M and Daubechies, Ingrid and Li, Jia and Postma, Eric and Wang, James Z},
  journal={IEEE Signal Processing Magazine},
  volume={25},
  number={4},
  pages={37--48},
  year={2008},
  publisher={IEEE}
}

@article{karayev2013recognizing,
  title={Recognizing image style},
  author={Karayev, Sergey and Trentacoste, Matthew and Han, Helen and Agarwala, Aseem and Darrell, Trevor and Hertzmann, Aaron and Winnemoeller, Holger},
  journal={arXiv preprint arXiv:1311.3715},
  year={2013}
}

@article{deussen2012feedback,
  title={Feedback-guided stroke placement for a painting machine},
  author={Deussen, Oliver and Lindemeier, Thomas and Pirk, S{\"o}ren and Tautzenberger, Mark},
  year={2012}
}

@article{tresset2013portrait,
  title={Portrait drawing by Paul the robot},
  author={Tresset, Patrick and Leymarie, Frederic Fol},
  journal={Computers \& Graphics},
  volume={37},
  number={5},
  pages={348--363},
  year={2013},
  publisher={Elsevier}
}

@inproceedings{hou2016patch,
  title={Patch-based convolutional neural network for whole slide tissue image classification},
  author={Hou, Le and Samaras, Dimitris and Kurc, Tahsin M and Gao, Yi and Davis, James E and Saltz, Joel H},
  booktitle={Proceedings of the IEEE conference on computer vision and pattern recognition},
  pages={2424--2433},
  year={2016}
}

@inproceedings{xu2014deep,
  title={Deep learning of feature representation with multiple instance learning for medical image analysis},
  author={Xu, Yan and Mo, Tao and Feng, Qiwei and Zhong, Peilin and Lai, Maode and Chang, Eric I-Chao},
  booktitle={2014 IEEE international conference on acoustics, speech and signal processing (ICASSP)},
  pages={1626--1630},
  year={2014},
  organization={IEEE}
}

@article{azimi2024patch,
  title={Patch-Based Oil Painting Forgery Detection Based on Brushstroke Analysis Using Generative Adversarial Networks and Depth Visualization},
  author={Azimi, Elhamsadat and Ashtari, Amirsaman and Ahn, Jaehong},
  journal={Applied Sciences},
  volume={15},
  number={1},
  pages={75},
  year={2024},
  publisher={MDPI}
}

@article{van2025patch,
  title={PATCH: a deep learning method to assess heterogeneity of artistic practice in historical paintings},
  author={Van Horn, Andrew and Smith, Lauryn and Mahmoud, Mahamad and McMaster, Michael and Pinchbeck, Clara and Martin, Ina and Lininger, Andrew and Ingrisano, Anthony and Lowe, Adam and Bayod, Carlos and others},
  journal={arXiv preprint arXiv:2502.01912},
  year={2025}
}

@book{zylinska2020ai,
  title={AI art: machine visions and warped dreams},
  author={Zylinska, Joanna},
  year={2020},
  publisher={Open humanities press}
}

@inproceedings{rossler2019faceforensics++,
  title={Faceforensics++: Learning to detect manipulated facial images},
  author={Rossler, Andreas and Cozzolino, Davide and Verdoliva, Luisa and Riess, Christian and Thies, Justus and Nie{\ss}ner, Matthias},
  booktitle={Proceedings of the IEEE/CVF international conference on computer vision},
  pages={1--11},
  year={2019}
}

@article{verdoliva2020media,
  title={Media forensics and deepfakes: an overview},
  author={Verdoliva, Luisa},
  journal={IEEE journal of selected topics in signal processing},
  volume={14},
  number={5},
  pages={910--932},
  year={2020},
  publisher={IEEE}
}

@inproceedings{davis2016empirically,
  title={Empirically studying participatory sense-making in abstract drawing with a co-creative cognitive agent},
  author={Davis, Nicholas and Hsiao, Chih-PIn and Yashraj Singh, Kunwar and Li, Lisa and Magerko, Brian},
  booktitle={Proceedings of the 21st international conference on intelligent user interfaces},
  pages={196--207},
  year={2016}
}

@article{schaldenbrand2024cofrida,
  title = {CoFRIDA: Self‑Supervised Fine‑Tuning for Human‑Robot Co‑Painting},
  author = {Schaldenbrand, Peter and Parmar, Gaurav and Zhu, Jun‑Yan and McCann, James and Oh, Jean},
  journal = {arXiv preprint arXiv:2402.13442},
  year = {2024},
  url = {https://arxiv.org/abs/2402.13442}
}

@inproceedings{schaldenbrand2022frida,
  title        = {FRIDA: A Collaborative Robot Painter with a Differentiable, Real2Sim2Real Planning Environment},
  author       = {Schaldenbrand, Peter and McCann, James and Oh, Jean},
  booktitle    = {IEEE International Conference on Robotics and Automation (ICRA)},
  year         = {2022},
}

@inproceedings{elgammal2018picasso,
  title={Picasso, matisse, or a fake? Automated analysis of drawings at the stroke level for attribution and authentication},
  author={Elgammal, Ahmed and Kang, Yan and Den Leeuw, Milko},
  booktitle={Proceedings of the AAAI Conference on Artificial Intelligence},
  volume={32},
  number={1},
  year={2018}
}

@inproceedings{ziaee2025fine,
  title={A Fine-Grained Artist Identification Method for Authentication and Attribution of Drawings Using Hatching Lines},
  author={Ziaee, Shahrzad and Elgammal, Ahmed and Mazzone, Marian},
  booktitle={2025 IEEE/CVF Conference on Computer Vision and Pattern Recognition Workshops (CVPRW)},
  pages={2153--2164},
  year={2025},
  organization={IEEE}
}

@article{sizyakin2020crack,
  title={Crack detection in paintings using convolutional neural networks},
  author={Sizyakin, Roman and Cornelis, Bruno and Meeus, Laurens and Dubois, Helene and Martens, Maximiliaan and Voronin, Viacheslav and others},
  journal={Ieee Access},
  volume={8},
  pages={74535--74552},
  year={2020},
  publisher={IEEE}
}

@article{shannon1948mathematical,
  title={A mathematical theory of communication},
  author={Shannon, Claude E},
  journal={The Bell system technical journal},
  volume={27},
  number={3},
  pages={379--423},
  year={1948},
  publisher={Nokia Bell Labs}
}

@article{goodall2012identification,
  title={Identification and authentication},
  author={Goodall, Rosemary A},
  year={2012}
}

@misc{hastie2009elements,
  title={The elements of statistical learning},
  author={Hastie, Trevor and Tibshirani, Robert and Friedman, Jerome and others},
  year={2009},
  publisher={Springer series in statistics New-York}
}

@book{spencer2004expert,
  title={The expert versus the object: judging fakes and false attributions in the visual arts},
  author={Spencer, Ronald D},
  year={2004},
  publisher={Oxford University Press}
}

@inproceedings{ojala1994performance,
  title={Performance evaluation of texture measures with classification based on Kullback discrimination of distributions},
  author={Ojala, Timo and Pietikainen, Matti and Harwood, David},
  booktitle={Proceedings of 12th international conference on pattern recognition},
  volume={1},
  pages={582--585},
  year={1994},
  organization={IEEE}
}

@article{breiman2001random,
  title={Random forests},
  author={Breiman, Leo},
  journal={Machine learning},
  volume={45},
  number={1},
  pages={5--32},
  year={2001},
  publisher={Springer}
}

@inproceedings{he2016deep,
  title={Deep residual learning for image recognition},
  author={He, Kaiming and Zhang, Xiangyu and Ren, Shaoqing and Sun, Jian},
  booktitle={Proceedings of the IEEE conference on computer vision and pattern recognition},
  pages={770--778},
  year={2016}
}

@article{oquab2023dinov2,
  title={Dinov2: Learning robust visual features without supervision},
  author={Oquab, Maxime and Darcet, Timoth{\'e}e and Moutakanni, Th{\'e}o and Vo, Huy and Szafraniec, Marc and Khalidov, Vasil and Fernandez, Pierre and Haziza, Daniel and Massa, Francisco and El-Nouby, Alaaeldin and others},
  journal={arXiv preprint arXiv:2304.07193},
  year={2023}
}

@article{cortes1995support,
  title={Support-vector networks},
  author={Cortes, Corinna and Vapnik, Vladimir},
  journal={Machine learning},
  volume={20},
  number={3},
  pages={273--297},
  year={1995},
  publisher={Springer}
}

@misc{artrecognition2024,
  title = {{Art Recognition}: {AI}-Based Art Authentication},
  author = {{Art Recognition}},
  year = {2024},
  howpublished = {\url{https://art-recognition.com/}},
  note = {Accessed: January 2025}
}

@article{cimpoi2015deep,
    title={Deep filter banks for texture recognition, description, and segmentation},
    author={Cimpoi, Mircea and Maji, Subhransu and Kokkinos, Iasonas and Vedaldi, Andrea},
    journal={International Journal of Computer Vision},
    volume={118},
    number={1},
    pages={65--94},
    year={2016}
}
\bibliographystyle{icml2026}

\newpage
\appendix
\onecolumn
\section{Appendix}

\begin{table}[htbp]
\centering
\caption{Per-fold cross-validation accuracies. Each fold holds out one painting for validation.}
\label{tab:per_fold_appendix}
\small
\begin{tabular}{clc}
\toprule
\textbf{Fold} & \textbf{Held-Out Painting} & \textbf{Accuracy (\%)} \\
\midrule
1  & human\_painting\_1 & 99.40 \\
2  & human\_painting\_2 & 90.26 \\
3  & human\_painting\_3 & 74.20 \\
4  & human\_painting\_4 & 69.00 \\
5  & human\_painting\_5 & 85.43 \\
6  & human\_painting\_6 & 96.97 \\
7  & human\_painting\_7 & 97.88 \\
8  & robot\_painting\_1 & 75.69 \\
9  & robot\_painting\_2 & 81.12 \\
10 & robot\_painting\_3 & 95.16 \\
11 & robot\_painting\_4 & 97.69 \\
12 & robot\_painting\_5 & 97.22 \\
13 & robot\_painting\_6 & 86.42 \\
14 & robot\_painting\_7 & 88.90 \\
15 & robot\_painting\_8 & 96.52 \\
\midrule
\multicolumn{2}{c}{\textbf{Mean ± Std Dev}} & \textbf{88.79 ± 10.41} \\
\bottomrule
\end{tabular}
\end{table}

\end{document}